\documentclass{article} % For LaTeX2e
\usepackage[final]{conference}

\usepackage{microtype}
\usepackage{hyperref}
\usepackage{url}
\usepackage{booktabs}
\usepackage{xspace}

\usepackage{lineno}

\usepackage{multirow}
\usepackage{makecell}
\usepackage{graphicx}
\usepackage{subcaption}
\definecolor{darkblue}{rgb}{0, 0, 0.5}
\hypersetup{colorlinks=true, citecolor=darkblue, linkcolor=darkblue, urlcolor=darkblue}

\newcommand{\pumba}{\textsc{PuMba}\xspace}

\title{Evaluating protein binding interfaces with \pumba}% Vision Mamba}

\author{Azam Shirali$^{1}$ \& Giri Narasimhan$^{1,2}$ \thanks{Corresponding Author.} \\
$^{1}$ Bioinformatics Research Group (BioRG),\\ Knight Foundation School of Computing and
Information Sciences, \\Florida International University; 11200 SW 8th 10 St, Miami, 33199,
USA. \\
$^{2}$ Biomolecular Sciences Institute, \\Florida International University; 11200 SW 8th St,
Miami, 33199, USA.
 \\
}

\begin{document}
\ifcolmsubmission
\linenumbers
\fi
\maketitle

\begin{abstract}
Protein–protein docking tools help in studying interactions between proteins, and are essential for drug, vaccine, and therapeutic development. However, the accuracy of a docking tool depends on a robust scoring function that can reliably differentiate between native and non-native complexes. 
PIsToN is a state-of-the-art deep learning–based scoring function that uses Vision Transformers in its architecture.
Recently, the Mamba architecture has demonstrated exceptional performance in both natural language processing and computer vision, often outperforming Transformer-based models in their domains. 
In this study, we introduce \pumba\  
(\textbf{P}rotein-protein interface
eval\textbf{u}ation with Vision \textbf{M}am\textbf{ba}), which improves PIsToN by replacing its Vision Transformer backbone with Vision Mamba. This change allows us to leverage Mamba’s efficient long-range sequence modeling for sequences of image patches. As a result, the model’s ability to capture both global and local patterns in protein–protein interface features is significantly improved. Evaluation on several widely-used, large-scale public datasets demonstrates that \pumba % our Vision Mamba-based PIsToN 
consistently outperforms its original Transformer-based predecessor, PIsToN.
\end{abstract}

\section{Introduction}
Protein-protein docking is crucial for understanding protein function, elucidating disease mechanisms, and guiding the development of therapies. 
A computational method for protein–protein docking generally involves two main steps. The first step is the sampling stage, where a large set of candidate conformations is generated. The second step is the scoring stage, during which a scoring function evaluates and ranks these conformations to identify the structure that most closely resembles the native protein complex. 
However, designing an efficient and reliable scoring function remains a major challenge \citep{shirali2025comprehensive}. 
Significant advancements have been achieved on addressing this problem, with deep learning (DL) methods frequently surpassing traditional physics-based and knowledge-based techniques. Among deep learning-based methods, PIsToN \citep{stebliankin2023evaluating} is a recent tool that demonstrates state-of-the-art performance in comparison to its competitors  \citep{shirali2025comprehensive}. 

PIsToN (evaluating \textbf{P}rotein binding \textbf{I}nterface\textbf{s} with \textbf{T}ransf\textbf{o}rmer \textbf{N}etworks) utilizes handcrafted chemical, physical, and geometric features extracted from the surfaces of the 3D structures of protein docking models. 
For each model, it computes the features representing them as a pair of surface patches—one from each interacting protein. Each patch is then transformed into a multi-channel image, with each channel representing a specific feature. The resulting pairs of images is then processed by a Vision Transformer (ViT) \citep{dosovitskiy2020image} to identify patterns that help distinguish native protein pairs from non-native decoys. The ViT embeddings are combined with "hybrid" empirical energy terms, allowing the model to capture the energetic characteristics of protein binding. A multi-attention mechanism then organizes the energy terms and interface features into five distinct groups, each processed by its own transformer network. Finally, the outputs of these networks are integrated through a transformer encoder, which generates the final classification score indicating whether the protein is native or non-native.

\textbf{Our main contributions:} In this work, we introduce \pumba (\textbf{P}rotein-protein interface eval\textbf{u}ation with \textbf{M}am\textbf{ba}), an enhanced version of PIsToN that replaces the vision transformer backbone with a Vision Mamba (ViM) architecture \citep{DBLP:conf/icml/ZhuL0W0W24}. ViM is based on the Mamba state space architecture \citep{gu2023mamba}, which enhances efficient modeling of long-range dependencies and bidirectional context propagation for sequences of image patches. By utilizing its selective state space mechanism, ViM can capture both local and global patterns in protein interface representations more efficiently than the attention-based model. Additional innovations in \pumba include  a reformulation to expose implicit attention matrices, thus imparting explainability features to \pumba. These advantages allow \pumba to learn discriminative features from multi-channel interface images more effectively, resulting in consistent improvements over the original PIsToN across standard classification metrics, as well as protein docking-specific evaluation measures. Our results demonstrate the potential of state space-driven architectures as a powerful alternative to transformers in modeling protein interactions.
By using different techniques, \pumba retains the powerful explainability features of PIsToN.

\section{Related work}
\label{gen_inst}

\subsection{Deep Learning in protein scoring functions}
Just as deep learning (DL) has achieved remarkable success in various applications, including text-related tasks (machine translation and sentiment analysis), image processing tasks (object detection and image segmentation), and graph analysis (social network analysis and molecular property prediction), it has also been applied to diverse problems in structural biology, including the design of powerful DL-based scoring functions for protein docking.

\citep{wang2021protein} introduced GNN-DOVE, an enhanced version of DOVE \citep{wang2020protein}, which utilizes two subgraphs to represent protein complex interfaces. Node features are encoded with one-hot vectors to capture properties like atom type, connection counts, and hydrogen atom numbers, while edges represent covalent and non-covalent bonds. A gate-augmented attention mechanism is used to identify atomic interaction patterns and assess their significance.

\citep{reau2023deeprank} introduced DeepRank-GNN, an improvement on DeepRank \citep{renaud2021deeprank}. This model represents protein-protein interfaces as residue-level graphs, incorporating features like residue type, charge, and buried surface area. It constructs two subgraphs: one for intra-protein residue connections and another for inter-protein residue connections. These are processed separately—one using a graph attention network (GAT) \citep{mahbub2022egret} and the other with an edge-aggregated graph attention network (EGRAT) \citep{mahbub2022egret}. The model calculates weighted contributions from neighboring residues to distinguish between native and non-native complexes.

\citep{sverrisson2021fast} developed dMaSIF, enhancing its predecessor, MaSIF \citep{gainza2020deciphering}, by representing protein surfaces as point clouds. Unlike MaSIF, dMaSIF assesses interactions between all atoms without precomputed features. It calculates ten geometric and six chemical features for each point on the interface, which are processed through (quasi-)geodesic convolutional layers to learn interaction patterns. dMaSIF generates fast and accurate binding scores.

AlphaFold3 (AF3) \citep{abramsonaccurate} significantly advanced biomolecular structure prediction. It modeled proteins, nucleic acids, small molecules, and ions, incorporating a scoring function directly into its prediction architecture for protein–protein complexes. Utilizing the Pairformer module, AF3 captured residue–residue relationships and employed a diffusion-based \citep{karras2022elucidating} refinement network for high-accuracy 3D structures. Key metrics included the predicted Template Modeling score (ipTM) \citep{xu2010significant} for interchain accuracy, predicted Template Modeling score (pTM) \citep{xu2010significant} for global structural accuracy, predicted Local Distance Difference Test (pLDDT) \citep{mariani2013lddt} for per-atom confidence, and Predicted Aligned Error (PAE) for positional uncertainty. These metrics allowed for effective ranking and evaluation of the predicted models.
For more details on comparing interface scoring methods, refer to the survey by  \citep{shirali2025comprehensive}.

\subsection{Deep Learning in machine vision}
Before the emergence of Transformers in vision tasks, convolutional neural networks (CNNs) were the dominant architecture, achieving significant success in areas such as image classification, object detection, and segmentation. CNNs excel at capturing local spatial patterns through hierarchical feature extraction, and their ability to learn translation-invariant features made them the preferred choice for most computer vision problems. 

However, the impressive success of transformers in natural language processing (NLP), driven by their self-attention mechanism and capability to model long-range dependencies, sparked interest in applying them to vision tasks. This interest led to the development of ViT \citep{dosovitskiy2020image}, which treats an image as a sequence of patches and utilizes transformer-based attention to capture global context. ViT quickly demonstrated competitiveness with, and often superiority over, CNNs, especially when large-scale training data was available, thanks to its flexibility in modeling complex relationships between image regions. Despite these advantages, transformers, including ViT, have inherent limitations, such as quadratic complexity in relation to sequence length (or the number of image patches), high memory requirements, and inefficiency with very large or high-resolution inputs. 

To address these challenges, Mamba \citep{gu2023mamba} was introduced, employing a selective state space model (SSM) that processes sequences with linear complexity, thus offering significantly greater scalability while maintaining strong modeling capacity. The success of Mamba in NLP tasks has inspired ViM \citep{DBLP:conf/icml/ZhuL0W0W24} to adapt these principles to image data. It does this by replacing the self-attention mechanism found in ViT-like architectures with efficient SSM-based blocks. This modification allows for faster training and enhances the handling of long or high-resolution sequences, all without sacrificing accuracy.

\subsection {Explainability and hidden attention in ViM}

``Explainability'' refers to the ability of a deep learning model to clearly demonstrate the reasoning behind its decisions. This makes it easier to understand why specific outcomes are predicted. In the context of transformers, these models rely on self-attention to capture the dependencies between different tokens. Self-attention enables the models to dynamically focus on different parts of the input sequence, assessing the relevance of each token in relation to others. This mechanism also allows the models to weigh the importance of tokens based on their contribution to the overall context, thus enhancing explainability as in \citep{chefer2021transformer} and \citep{abnar2020quantifying}.

Unlike transformers, which provide explicit attention matrices that can be used for explainability, Mamba models lack such readily accessible structures, creating a gap in their interpretability. This lack of dedicated tools complicates debugging Mamba models or applying them confidently in sensitive domains where explainability is crucial.

\citep{DBLP:journals/corr/abs-2403-01590} reveals that SSM models like Mamba can be reformulated to expose implicit attention matrices. This effectively demonstrates that Mamba implements a variant of causal self-attention. In contrast to transformers, where each attention head produces a single attention matrix, a single Mamba layer generates significantly more attention matrices due to its multiple channels, allowing for finer-grained tracking of token-to-token influence. The authors show that while transformer attention relies on explicit dot-product interactions between queries and keys, Mamba’s hidden attention modulates influence over time, leading to more efficient modeling of long-range dependencies.

\section{Our approach}
\label{headings}

Building on PIsToN \citep{stebliankin2023evaluating}, our model (\pumba) replaces the ViT backbone with ViM, an SSM–based architecture introduced for visual sequence modeling. Similar to PIsToN, our approach is trained to distinguish native-like docking models from non-native conformations by learning from the surface characteristics of binding interfaces. 

Given the PDB format \citep{berman2000protein} of a protein–protein complex, the structure is first refined using FireDock \citep{andrusier2007firedock}, which also computes binding free energy components, including Van der Waals, desolvation, insideness, hydrogen bonds, disulfide bonds, electrostatics, $\pi$-stacking, cation-$\pi$ and aliphatic interactions \citep{stebliankin2023evaluating}. The refined complex is then cropped within a distance $r_{\textrm{surf}}$ from the interaction center, defined as the geometric center of all atoms located within 5{\AA} of any atom in the binding partner. The solvent-excluded surface is subsequently triangulated and rescaled to 1{\AA} granularity using the MaSIF data preparation module \citep{gainza2020deciphering}. Surface features are then computed over circular patches on the interacting regions, where a patch consists of surface vertices within geodesic distance $r_{\textrm{surf}}$ from the interaction center. For every surface point, six features are extracted which are shape index, curvature, hydrogen bonding potential, charge, hydropathy, and Reletive Accessible Surface Area (RASA). Additionally, a ``patch dist'' feature is derived from the Euclidean distances between corresponding points on paired patches. RASA values are obtained using DSSP v2.3 \citep{touw2015series}. All computed features are converted into multi-channel 2D images by projecting patch surface points onto a plane using multidimensional scaling (MDS) \citep{mead1992review}, producing a $2r_{\textrm{surf}} \times 2r_{\textrm{surf}}$ grid for each feature channel, with pixel intensities smoothed over nearest neighbors ( see \citep{stebliankin2023evaluating} for details). These multi-channel images are then processed by a ViM backbone, which treats sequences of image patches as ordered tokens, enabling efficient modeling of both local and long-range dependencies across surface features.

\subsection{\pumba architecture}

The multi-channel images obtained from the binding interface are provided as inputs to a ViM network. The ViM architecture follows the introduced in \citep{DBLP:conf/icml/ZhuL0W0W24}, with modifications to accommodate $N$ input channels corresponding to the number of computed surface features, rather that the conventional RGB channels. Each image is subdevided into square patches of size $l$, producing $P=(a/l)^2$ mini-patches, where $a$ denotes the image size (i.e. the length of each side in pixels). These patches are then linearly projected into $M$-dimensional token embeddings \citep{stebliankin2023evaluating}. Following \citep{DBLP:conf/icml/ZhuL0W0W24}, a class token is inserted at the center of the embedding sequence, as this position has been shown to yield optimal performance. The output provides a ranking score for each protein docking model, distinguishing between native-like and non-native conformations.

\begin{figure}[t]
    \centering
    \includegraphics[width=0.95\linewidth]{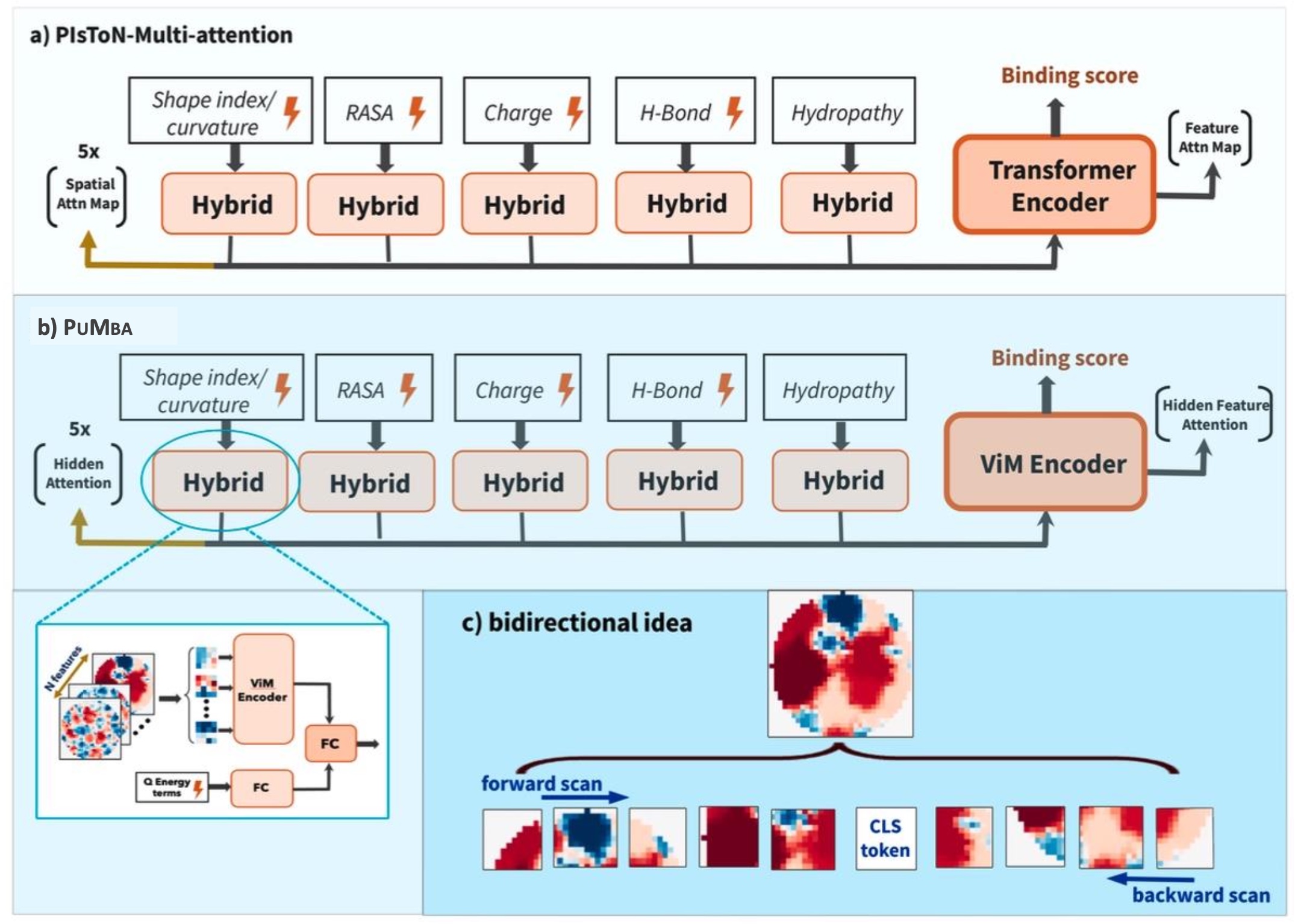}
    \begin{subfigure}[t]{\linewidth}
        \centering
        \phantomcaption
        \label{fig:arch_a}
    \end{subfigure}
    \begin{subfigure}[t]{\linewidth}
        \centering
        \phantomcaption
        \label{fig:arch_b}
    \end{subfigure}
    \begin{subfigure}[t]{\linewidth}
        \centering
        \phantomcaption
        \label{fig:arch_c}
    \end{subfigure}
    \caption{Overview of PIsToN and \pumba architectures: 
    (\subref{fig:arch_a}, \subref{fig:arch_b}) Multi-attention architecture with hybrid feature processing in PIsToN and \pumba, 
    (\subref{fig:arch_c}) bidirectional scans of patches of each feature in forward and backward SSM blocks in \pumba. The CLS token is inserted into the middle of the patch sequence to maximize bidirectional context utilization.}
    \label{fig:arch}
\end{figure}

As in PIsToN (Fig.\ref{fig:arch_a}), related surface features and energy terms are organized into five functional groups: shape, RASA, charge, hydrogen bonds, and hydropathy. In \pumba (Fig.\ref{fig:arch_b}), each group is processed by an independent hybrid module in which the original ViT encoder from PIsToN is replace with a ViM encoder, followed by a fully connected integration layer. The resulting representations are then aggregated through a final ViM encoder to generate a binding score for each input protein docking model. 

The challenge with image-like protein surface patches is that they are inherently non-sequential. In attention-based models like ViT, the attention mechanism can directly access all tokens simultaneously, making spatial order less critical as long as positional encodings are included. In contrast, ViM is an SSM-based model that updates its hidden state recurrently based on both the previous state and the current input. This sequential nature makes the input order important; early states carry less contextual information than later ones when processed in a single forward pass. To address this challenge in vision tasks, ViM utilizes Bidirectional SSMs idea, which consists of two separate SSMs within each encoder block: the Forward and Backward SSM to process the patch embeddings in the forward and reverse orders. This approach effectively allows future patch information to be passed backward through the model. The outputs of these two SSMs are fused, enabling each token to access both past and future contexts.

As suggested in the ViM architecture, in \pumba we added a class token as an additional token positioned in the middle to enhance classification performance. 
% We adopt this approach by placing the class token in the middle of patch token embeddings. 
Fig.\ref{fig:arch_c} illustrates an example of patch embedding, where the image of the hydropathy feature is split into small patches with a class token placed in the middle. In \pumba, the internal operations of each encoder layer—including SSM computation, bidirectional fusion, and positional encoding—are consistent with those described in the ViM paper \citep{DBLP:conf/icml/ZhuL0W0W24}.

To incorporate the explainability features of PIsToN, we follow the framework established in \citet{DBLP:journals/corr/abs-2403-01590}. This ensures that the important feature of explainability is not sacrificed by replacing the transformer-based model with a Mamba-based model. As in PIsToN, our model offers a two-stage attention interpretation: at the token level within each branch, it captures the token-to-token dependency dynamics across the intermediate Mamba layers, identifying which pixels the model pays more attention to for each feature. At the feature level across branches (in the final ViM encoder), it determines which feature(s) contribute most significantly to the final decision. An example of this interpretation is provided in Section \ref{results}.

We follow the training procedure described by \citet{stebliankin2023evaluating} for PIsToN, adopting the same training and validation datasets. The loss formulation remains unchanged, combining binary cross-entropy, supervised contrastive loss \citep{khosla2020supervised}, and margin ranking loss \citep{chen2019looks}. The training strategy is also replicated, including input preprocessing, optimizer choice, and learning rate schedule, to ensure consistency with their reported performance (The AdamW optimizer with the learning rate of 0.0001 and weight decay of 0.001 for weights optimization). For a detailed description of these procedures, readers are referred to \citet{stebliankin2023evaluating}.

\section{Experimental Results}\label{results}
To evaluate the effectiveness of the proposed \pumba architecture, we conducted a series of experiments comparing its performance to the original PIsToN model using popular and well-known datasets, which we describe in this section. We employed all eight datasets reviewed in the survey by \citep{shirali2025comprehensive} to ensure a broad and diverse evaluation of PIsToN and \pumba. These datasets span widely used benchmark collections, covering a variety of docking scenarios and difficulty levels. %Table \ref{tab:datasets} presents the size and the number of positive and negative protein docking models in each dataset.% 
More details about each dataset can be found in \citep{shirali2025comprehensive}. Training and testing were performed on a GeForce GTX 1080 Ti GPU with 8 GB memory, 256 GB RAM, and a 28-core Intel Xeon E5-2650 CPU.

Following the evaluation process suggested by \citep{shirali2025comprehensive}, we evaluated \pumba against its predecessor PIsToN across the same eight widely used protein–protein docking benchmark datasets: CAPRI Score v2022 (Difficult/Easy), CAPRI Score Refined, BM4, BM5, Dockground, PDB 2023, and MaSIF test. The evaluation considered several standard metrics, including the area under the ROC curve (AUC ROC), average precision (AP), balanced accuracy (BA), F1 score, precision, and recall. These metrics collectively assess the models’ ability to rank native over non-native docking models.

Table \ref{tab:class metrics} demonstrates that \pumba consistently matched or outperformed PIsToN across various datasets, with particularly notable gains in the challenging datasets. For instance, in the CAPRI v2022 Difficult dataset, AUC ROC increased from 64.77 to 68.83, AP rose from 3.02 to 10.49, and both precision and recall improved, with precision increasing from 3.47 to 9.07 and recall increasing from 70.49 to 76.52. In Dockground dataset, PuMba achieved a more than two-fold improvement in AP (from 10.71 to 25.33) along with a higher F1 score (2.58 increase). Even in datasets where PIsToN already performed at a high level, such as BM5 and PDB 2023, \pumba maintained or slightly surpassed the original performance, showing robustness without compromising accuracy.

The success rate evaluation in Table \ref{table:succ_rate_results} further supports these improvements. Success rate indicates how frequently a docking method yields at least one model of acceptable quality among its top 1, top 10, top 25, top 100, and top 200 predicted docking models. A higher success rate directly enhances drug discovery by increasing early enrichment, decreasing the number of compounds that need testing, and accelerating design-make-test cycles. Table \ref{table:succ_rate_results} shows that in CAPRI v2022 Difficult, the Top1 success rate increased from 17\% to 22\%, and in Dockground, it rose from 15\% to 32\%. Gains were also evident in BM4, where the Top1 rate improved from 54\% to 57\%, and in CAPRI v2022 Easy, where it increased from 41\% to 48\%. At higher ranking thresholds (Top100 and Top200), \pumba achieved perfect or near-perfect success rates in several datasets, further demonstrating its capacity to prioritize native-like docking models effectively.

To evaluate the quality of the protein docking models scored by each scoring function, they can be classified into four categories: incorrect, acceptable, medium, or high. This classification is based on the CAPRI (Critical Assessment of Predicted Interactions) criteria \citep{lensink2007docking} and is performed using the CAPRI-Q tool \citep{collins2024capri}. Table \ref{tab:capri_quality} shows the ability of  PIsToN and \pumba in ranking docking models of varying quality among different top-k predictions (k = 1, 10, or 100). Across all eight benchmark datasets, \pumba consistently identifies a greater number of acceptable-to-high-quality models than PIsToN, particularly within the top 10 and top 100 predictions. For example, in the BM5 dataset, it retrieved 44 acceptable-quality docking models in the top 10 and 78 in the top 100, compared to PIsToN’s 40 and 70. In the CAPRI Score v2022 (Easy), \pumba found 92 acceptable models in the top 100, exceeding PIsToN’s 66. Notably, in PDB 2023, it recovered 98 high-quality docking models in the top 100, while PIsToN retrieved 96, also showing better performance in medium and acceptable categories. Overall, the models ranked by \pumba are of higher quality than those by PIsToN.

\begin{table}[t]
\begin{center}
\small
% {\fontsize{8.5pt}\selectfont
\begin{tabular}{llcccccc}
\toprule
\textbf{Dataset} & \textbf{Method} & \textbf{AUC ROC} & \textbf{AP} & \textbf{BA} & \textbf{F1} & \textbf{Precision} & \textbf{Recall} \\
\midrule

\textbf{CAPRI Score v2022} & PIsToN & 64.77 & 3.02 & \textbf{63.05} & 66.16 & 3.47 & 70.49 \\
\textbf{(Difficult)} & \pumba & \textbf{68.83} & \textbf{10.49} & 62.71 & \textbf{67.23} & \textbf{9.07} & \textbf{76.52} \\
\midrule

\textbf{CAPRI Score v2022} & PIsToN & \textbf{80.15} & 46.24 & \textbf{73.08} & 60.51 & 55.89 & 65.95 \\
\textbf{(Easy)} & \pumba & 79.91 & \textbf{48.15} & 72.12 & \textbf{63.46} & \textbf{58.19} & \textbf{69.80} \\
\midrule

\textbf{CAPRI Score Refined} & PIsToN & \textbf{81.25} & 24.86 & 73.24 & \textbf{42.21} & \textbf{30.37} & 69.16 \\
\textbf{ } & \pumba & 79.95 & \textbf{25.19} & \textbf{74.21} & 39.38 & 29.15 & \textbf{76.56} \\
\midrule

\textbf{BM4} & PIsToN & 70.84 & 36.46 & 62.71 & \textbf{43.67} & \textbf{53.75} & 36.77 \\
\textbf{ } & \pumba & \textbf{72.50} & \textbf{37.27} & \textbf{64.64} & 43.04 & 52.27 & \textbf{45.13} \\
\midrule

\textbf{BM5}& PIsToN & 91.84 & \textbf{23.64} & \textbf{74.90} & 45.14 & 38.16 & 55.24 \\
\textbf{ } & \pumba & \textbf{92.02} & 22.31  & 72.74 & \textbf{46.86}  & \textbf{41.10}  & \textbf{56.49} \\
\midrule

\textbf{Dockground} & PIsToN & 72.04 & 10.71 & 60.40 & 22.13 & 15.62 & 37.92 \\
\textbf{ } & \pumba & \textbf{72.94} & \textbf{25.33}  & \textbf{64.12} & \textbf{24.71}  & \textbf{18.15}  & \textbf{39.46} \\
\midrule

\textbf{PDB 2023} & PIsToN & 93.86 & 45.85 & 78.21 & 66.05 & \textbf{78.89} & 56.80 \\
\textbf{ } & \pumba & \textbf{93.91} & \textbf{46.11}  & \textbf{80.23} & \textbf{66.72}  & 75.13  & \textbf{58.66} \\
\midrule

\textbf{MaSIF test} & PIsToN & \textbf{93.55} & 81.08 & 85.43 & 85.97 & \textbf{87.72} & 82.40 \\
\textbf{ } & \pumba & 93.42 & \textbf{81.45} & \textbf{85.97} & \textbf{86.34} & 87.69 & \textbf{84.12} \\
\bottomrule

\end{tabular}
\end{center}

\tiny{\textcolor{white}{----}Bold values indicate the best value for each column.} 
\caption{Evaluation of PIsToN and \pumba across eight benchmark datasets.}
\label{tab:class metrics}
\end{table}

\begin{table}[t]
\begin{center}
\small
% \scriptsize
% {\fontsize{8.5pt}\selectfont
\begin{tabular}{llccccc}
\toprule
\textbf{Dataset} & \textbf{Method} & \textbf{Top1} & \textbf{Top10} & \textbf{Top25} & \textbf{Top100} & \textbf{Top200} \\
\midrule

\textbf{CAPRI v2022 (Difficult)} & PIsToN & 17 & 28 & 46 & \textbf{71} & 78 \\
& \pumba    & \textbf{22} & \textbf{31} & \textbf{47} & 69 & \textbf{85} \\
\midrule

\textbf{CAPRI v2022 (Easy)} & PIsToN & 41 & 82 & \textbf{87} & 94 & 97 \\
      & \pumba    & \textbf{48} & \textbf{83} & \textbf{87} & \textbf{96} & \textbf{100} \\
\midrule

\textbf{CAPRI Refined} & PIsToN & 38 & \textbf{69} & \textbf{76} & 76 & \textbf{100} \\
         & \pumba & \textbf{40} & \textbf{69} & 75 & \textbf{81} & \textbf{100} \\
\midrule

\textbf{BM4} & PIsToN & 54 & 89 & 89 & \textbf{94} & \textbf{100} \\
                  & \pumba & \textbf{57} & \textbf{90} & \textbf{90} & 92 & \textbf{100} \\
\midrule

\textbf{BM5} & PIsToN & \textbf{66} & 93 & \textbf{100} & \textbf{100} & \textbf{100} \\
                    & \pumba & 64 & \textbf{95} & \textbf{100} & \textbf{100} & \textbf{100} \\
\midrule

\textbf{Dockground} & PIsToN & 15 & 55 & 81 & \textbf{100} & \textbf{100} \\
            & \pumba & \textbf{32} & \textbf{59} & \textbf{83} & \textbf{100} & \textbf{100} \\
\midrule

\textbf{PDB 2023} & PIsToN & 88 & 96 & 98 & \textbf{100} & \textbf{100} \\
               & \pumba & \textbf{89} & \textbf{99} & \textbf{99} & \textbf{100} & \textbf{100} \\
\bottomrule

\end{tabular}
% \\[0.3em] % small space
% \tiny{Bold values indicate the best value for each column.}
\end{center}
% \normalsize
\tiny{\textcolor{white}{---   ---  ---- --    --}Bold values indicate the best value for each column.} 
\caption{Success rates (\%) of PIsToN and \pumba on eight different datasets.}
\label{table:succ_rate_results}

\end{table}

\begin{table}[t]
\begin{center}
\small
% \begin{tabular}{llccc}
\setlength{\tabcolsep}{4pt}
\begin{tabular}{ll*{9}{c}}
\toprule
\textbf{Dataset} & \textbf{Method} & \multicolumn{3}{c}{\textbf{acceptable}} & \multicolumn{3}{c}{\textbf{medium}} & \multicolumn{3}{c}{\textbf{high}} \\
\cmidrule(lr){3-5} \cmidrule(lr){6-8} \cmidrule(lr){9-11}
 & & \textbf{top1} & \textbf{top10} & \textbf{top100} & \textbf{top1} & \textbf{top10} & \textbf{top100} & \textbf{top1} & \textbf{top10} & \textbf{top100} \\
\midrule

\textbf{CAPRI Score v2022}   
 & PIsToN & 17 & 28 & 67 & 0 & 3 & \textbf{35} & 0 & 0 & 0 \\
  \textbf{(Difficult)} & \pumba & \textbf{19} & \textbf{32} & \textbf{71} & \textbf{2} & \textbf{6} & 33 & 0 & 0 & \textbf{3} \\
\midrule

\textbf{CAPRI Score v2022}   
 & PIsToN & \textbf{17} & 51 & 84 & 20 & \textbf{66} & 87 & 2 & 20 & 30 \\
 \textbf{(Easy)} & \pumba & 16 & \textbf{53} & \textbf{92} & \textbf{24} & 65 & \textbf{89} & \textbf{5} & \textbf{23} & \textbf{37} \\
\midrule
\textbf{CAPRI Scores}   
 & PIsToN & 7 & \textbf{46} & 61 & 30 & \textbf{46} & 53 & 0 & \textbf{7} & 23 \\
 \textbf{Refined} & \pumba & \textbf{12} & 45 & \textbf{63} & \textbf{32} & \textbf{46} & \textbf{57} & 0 & \textbf{7} & \textbf{26} \\
\midrule
\multirow{2}{*}{\textbf{BM4}} 
 & PIsToN & 5 & 26 & \textbf{42} & 5 & 26 & \textbf{42} & 5 & 21 & 42 \\
 & \pumba & \textbf{15} & \textbf{33} & 41 & \textbf{11} & \textbf{28} & 40 & \textbf{6} & \textbf{23} & \textbf{43} \\
\midrule
\multirow{2}{*}{\textbf{BM5}} 
 & PIsToN & \textbf{6} & 40 & \textbf{80} & \textbf{40} & 60 & 66 & \textbf{20} & 73 & \textbf{100} \\
 & \pumba & 5 & \textbf{44} & 78 & 38 & \textbf{64} & \textbf{72} & 18 & \textbf{77} & \textbf{100} \\
 \midrule
 
 \multirow{2}{*}{\textbf{Dockground}} 
 & PIsToN & 0 & 1 & \textbf{28} & 24 & 71 & \textbf{98} & 1 & 8 & 17 \\
 & \pumba & \textbf{1} & \textbf{4} & 22 & \textbf{31} & \textbf{75} & \textbf{98} & \textbf{6} & \textbf{15} & \textbf{31} \\
\midrule
\multirow{2}{*}{\makecell{\textbf{PDB 2023} \\ }}  
 & PIsToN & 1 & \textbf{44} & \textbf{48} & 9 & 30 & 30 & \textbf{76} & 92 & 96 \\
 & \pumba & \textbf{2} & 36 & \textbf{48} & \textbf{13} & \textbf{33} & \textbf{41} & \textbf{76} & \textbf{95} & \textbf{98} \\
 
\bottomrule
% \text{\tiny{Bold values indicate the best value for each column.}} 
\end{tabular}
\end{center}
\tiny{Bold values indicate the best value for each column.}
\caption{Quality of docking models scored by PIsToN and \pumba across eight benchmark datasets. Values represent the number of docking models classified into each CAPRI quality category among the top 1, 10, and 100 predictions}
\label{tab:capri_quality}
\end{table}

\begin{figure}[t]
\begin{center}
\includegraphics[width=0.8\linewidth]{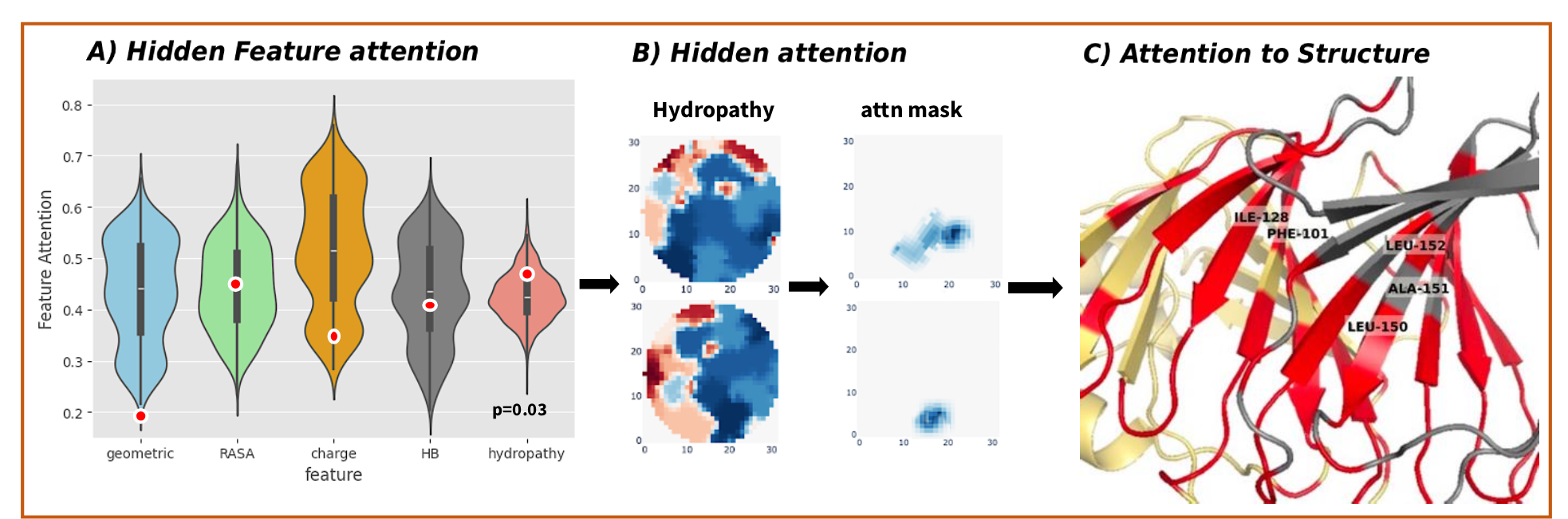}
\end{center}
\caption{Explainability in \pumba with homodimer (PDB ID 1J3R, chains A and B).} \label{fig:XAI}
\end{figure}

To assess the ability of \pumba to be interpretable and the fair comparison, we followed the explainability protocol of PIsToN while adapting it to \pumba’s hidden attention values. For each protein complex, two sets of attention maps were extracted per feature branch (Fig.\ref{fig:arch_b}): hidden attention maps and hidden feature attention maps from the final binding score module. These maps were averaged across all heads and layers, and the z-score was calculated to highlight the most influential pixels and regions (z-score$\geq$-1.96, p$\leq$0.05). The resulting high-importance regions were overlaid onto the corresponding feature images to visualize feature(s) contributed most to the predicted score. Fig.\ref{fig:XAI} illustrates an example of the explainability of \pumba, which is consistent with the PIsToN model used. Notably, \pumba focused on the same features as PIsToN when deciding about classifying complexes as native or non-native conformations, demonstrating that the model maintains strong explainability while delivering improved predictive accuracy. For this example, both models placed greater attention on the hydropathy feature when predicting a native protein complex. Furthermore, by mapping the high-importance interface pixels back to the original structure in PyMOL \citep{schrodinger2015pymol}, we observed that the residues highlighted for the hydropathy feature were predominantly hydrophobic, confirming the correctness of our model’s attention and showing that its focus aligns precisely with the biologically relevant interaction regions.

\section{Conclusions}
We introduce \pumba, a scoring tool designed to rank protein docking models. The ViT backbone of PIsToN was replaced with an efficient, bidirectional ViM architecture in \pumba. By utilizing selective state space modeling, \pumba effectively captures both global and local binding interface patterns. Hidden attention analysis further supports interpretability, even in the absence of explicit attention maps. % By retaining PIsToN’s original training pipeline, we ensure a fair comparison between the two methods. 
Experimental results show that \pumba consistently outperforms PIsToN across standard classification metrics and the CAPRI docking quality criteria. These findings underscore the potential of Mamba-based architectures to advance the assessment of biomolecular docking.

\section*{Acknowledgments}

Our sincere thanks to the designers of PIsToN for sharing their insights on this problem.

\section*{Data and code availability}
Datasets and code are available at: {\url{https://zenodo.org/records/12681335} and \url{https://github.com/Azam-Shi/PuMba}, respectively.
% \pumba code is available at: 

\bibliography{colm2025_conference}
\bibliographystyle{colm2025_conference}

\end{document}